\documentclass[letterpaper, 10 pt, conference]{ieeeconf}  % Comment this line out if you need a4paper

\IEEEoverridecommandlockouts                              % This command is only needed if 
\usepackage{amsmath}
\usepackage{amsfonts}
\usepackage{graphicx}
\usepackage{breqn}
\usepackage{subcaption}
\usepackage{algorithm}
\usepackage{algpseudocode}

\usepackage[dvipsnames]{xcolor}
\usepackage[deletedmarkup=sout,authormarkup=superscript]{changes}
\definechangesauthor[name={Frank}, color=Purple]{FP}

\newtheorem{theorem}{Theorem}[section]
\newtheorem{lemma}[theorem]{Lemma}

\newcommand{\FigureFolder}{.}

\title{\LARGE \bf
Convex synthesis and verification of control-Lyapunov and barrier functions with input constraints
}

\author{Hongkai Dai$^{1}$ and Frank Permenter$^{1}$% <-this % stops a space
\thanks{$^{1}$Toyota Research Institute, US
        {\tt\small hongkai.dai, frank.permenter@tri.global}}%
}

\begin{document}

\maketitle
\thispagestyle{empty}
\pagestyle{empty}

%%%%%%%%%%%%%%%%%%%%%%%%%%%%%%%%%%%%%%%%%%%%%%%%%%%%%%%%%%%%%%%%%%%%%%%%%%%%%%%%
\begin{abstract}
        Control Lyapunov functions (CLFs) and control barrier functions (CBFs)
        are widely used tools for synthesizing controllers subject to stability
        and safety constraints.  Paired with online optimization, they  provide
        stabilizing control actions that satisfy input constraints and avoid unsafe
        regions of state-space.  Designing CLFs and CBFs with rigorous
        performance guarantees is computationally challenging. To certify
        existence of control actions, current techniques not only design a
        CLF/CBF, but also a nominal controller.  This can make the synthesis task more expensive,
        and performance estimation more conservative. 
        In this work, 
        we characterize polynomial CLFs/CBFs using sum-of-squares conditions,
        which can be directly certified using convex optimization.
        This yields a CLF and CBF synthesis technique that does not rely 
        on a nominal controller.  We then present 
        algorithms for iteratively enlarging estimates
        of the stabilizable and safe regions. We demonstrate our algorithms on a
        2D toy system, a pendulum and a quadrotor.
\end{abstract}

\section{INTRODUCTION}
When synthesizing controllers for dynamical systems, it is often paramount to
ensure that the closed-loop system 
always converges to the desired state and always avoids unsafe regions.
These goals can be met by using control Lyapunov functions (CLFs)~\cite{sontag1983lyapunov} 
for stability, and control barrier functions (CBFs)~\cite{ames2016} for safety. CLFs and CBFs are designed
such that their online minimization leads to desired
performance goals.  Local verification  amounts
to certifying these minimization problems are feasible 
on some subset of state-space, which can be a challenging computational task. 

CLFs and CBFs have been used extensively in controller design for various
applications, including legged locomotion \cite{galloway2015torque,
grandia2020, nguyen2020dynamic}, autonomous driving \cite{ames2014control,
chen2017obstacle, zeng2021safety}, and robot arm manipulation
\cite{murtaza2022safety}. Recently, they have been used to guide 
learning-based methods \cite{cheng2019, choi2020reinforcement, kang2022lyapunov}. 
The CLFs/CBFs are typically synthesized by hand or learned from data
\cite{robey2020learning, dawson2022safe} without rigorous verification.
In this paper, we provide a new synthesis technique with formal guarantees
that is conceptually simpler than previous formal methods \cite{jarvis2003, majumdar2013control, ames2019control, wang2022safety}.
In particular, our technique allows one to verify
CLFs and CBFs by solving a single convex optimization problem,
under the assumption the dynamics are control-affine and
the input constraints are polyhedral (for example, robots subject to torque limits for each motor).  This in turn
leads to synthesis algorithms based on sequential convex
optimization.

\begin{figure}
	\centering
	\includegraphics[width=0.4\textwidth]{\FigureFolder/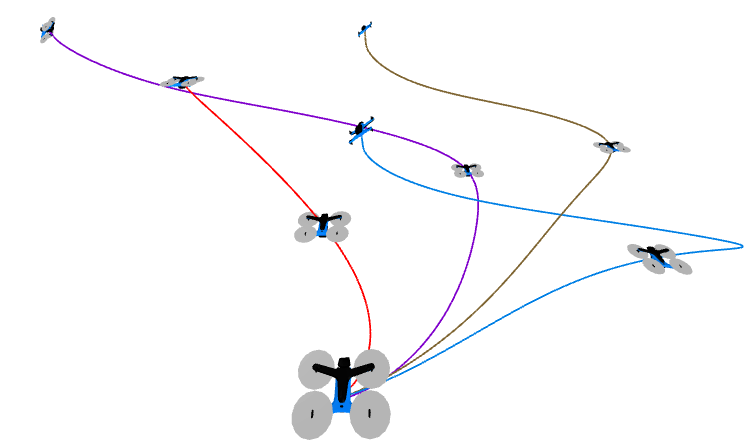}
	\caption{With our certified CLF, the QP controller can stabilize the quadrotor from many distant states (including ones that are 10 meters away or the initial roll angle of $140^\circ$) to the desired hovering state.}
	\label{fig:quadrotor_clf_traj}
	\vspace{-12pt}
\end{figure}

The basis of our technique is Sum-Of-Squares (SOS) optimization \cite{blekherman2012, parrilo2000},
a widely used tool for controller synthesis and verification, including CLF and CBF design \cite{tan2004, clark2021verification, jarvis2003, majumdar2013control, ames2019control, wang2022safety}.
These previous methods either ignore input constraints \cite{tan2004, clark2021verification}
or rely on the joint synthesis of a nominal control law of polynomial form \cite{jarvis2003, majumdar2013control, ames2019control, wang2022safety}.
Reliance on this polynomial controller is restrictive if the actual stable/safe
control policy is not polynomial. It also limits scalability, as the
size of the SOS program increases rapidly with the degree of the controller. 

In this paper, we derive new necessary and sufficient conditions for CLFs/CBFs
for polynomial dynamical systems with input constraints. We formulate these
conditions as SOS feasibility problems, and present an iterative algorithm
for enlarging an inner approximation of the stabilizable and/or safe region
via sequential SOS optimization. We demonstrate our algorithm on different systems, including a
2D toy example, an inverted pendulum and a quadrotor. To our best knowledge
this is the first formal method for CLF/CBF synthesis that both accounts
for input constraints and explicitly avoids construction of a nominal 
controller.

\section{BACKGROUND}
\label{sec:background}
In this section we give a brief introduction to Sum-Of-Squares (SOS)
techniques for certifying polynomial non-negativity.  
To begin, a polynomial $p(x)$ is a sum-of-squares (sos) iff $p(x) = \sum_i q_i(x)^2$ for some polynomials $q_i(x)$. Clearly $p(x)$ being sos implies that $p(x)\ge 0 \,\forall x$. If $p(x)$ has degree $2d$, then it is an sos polynomial if and only if
\begin{align}
   p(x) = m(x)^TS\,m(x), S\succeq 0 \label{eq:sos},
\end{align}
where $m(x)$ is a vector consisting of all monomials of degree at most $d$. Given $p(x)$ and $m(x)$, existence of $S$ can be checked
using semidefinite programming~\cite{parrilo2000, blekherman2012}.

Sum-of-squares optimization can also certify polynomial non-negativity on the
feasible set of finitely-many polynomial inequalities, i.e.,
on a semialgebraic set $K$ of the form
$\{x \in \mathbb{R}^n |b_1(x)\ge 0,\hdots,b_m(x)\ge0\}$.
The underlying certificates 
employ the \textbf{preorder} of $b_i$, defined as 
\begin{dmath}
		\textbf{preorder}(b_1(x),\hdots, b_m(x)) =
		\left\{l_0(x) + \sum_{i=1}^m l_i(x)b_i(x) +\sum_{i\neq j} l_{ij}(x)b_i(x)b_j(x)
		+ \sum_{i\neq j \neq k} l_{ijk}(x)b_i(x)b_j(x)b_k(x) + \hdots
		|\; l_0(x), l_i(x), l_{ij}(x), l_{ijk}(x),\hdots \text{ are all sos}\right\}.
		\label{eq:preorder}
\end{dmath}

For a given polynomial $p(x)$, the \textit{Positivstellensatz}
\cite{stengle1974}\cite[Section~3.6]{laurent2009} states that
\begin{equation}
\begin{gathered}
	p(x)\ge 0 \text{ on } K\\
	\Updownarrow\\
	\exists q(x), r(x)\in\textbf{preorder}(b_1(x),\hdots,b_m(x)),k\in\mathbb{N}\\
	\text{ s.t } p(x)q(x) = p(x)^{2k} + r(x).
\end{gathered}
\end{equation}
In other words, if $p(x)$ is non-negative on $K$, then there is a certificate
of this fact defined by polynomials $q(x)$ and $r(x)$ in the preorder.
Further, for fixed $k$, finding this certificate can be cast as a semidefinite
program~\cite{blekherman2012}. 

Unfortunately, a generic element of the preorder is the summation of $2^m$
different polynomials each scaled by a different sum-of-squares polynomial.
This exponential complexity motivates simpler (sufficient) conditions for
non-negativity. A common simplification---called the \emph{S-procedure}~\cite{parrilo2000}---is existence of polynomials $\bar{r}_i(x), i=0,\hdots,m$ satisfying
\begin{subequations}
\begin{align}
	(1+\bar{r}_0(x))p(x)-\sum_{i=1}^m\bar{r}_i(x)b_i(x)\text{ is sos}\label{eq:s_procedure1}\\
	\bar{r}_i(x)\text{ is sos}, i=0,\hdots,m.
\end{align}\label{eq:s_procedure}
\end{subequations}
The equation~\eqref{eq:s_procedure1} implies that $p(x)\ge 0$ on $K$ because, by definition, 
$\bar{r}_i(x) b_i(x) \ge 0$ on $K$. Frequently, we simplify this condition further by taking $\bar{r}_0(x)=0$.

\section{Problem formulation}\label{sec:problem_formulation}
We consider a control-affine dynamical system of the form
\begin{align}
	\dot{x} = f(x) + g(x)u, \;\; u \in \mathcal{U},
\end{align}
where $x\in\mathbb{R}^{n_x}$ and $u\in\mathbb{R}^{n_u}$ denote the state
and control,
$f : \mathbb{R}^{n_x} \rightarrow \mathbb{R}^{n_x}$  and $g : \mathbb{R}^{n_x}
\rightarrow \mathbb{R}^{n_x \times n_u}$ are polynomial functions of $x$
and $\mathcal{U} \subset \mathbb{R}^{n_u}$ denotes the set of admissible inputs,
which we assume to be a convex polytope. Equivalently, we assume existence of a finite
set of points $u^i$ satisfying
\begin{align}
  \mathcal{U} = \text{ConvexHull}(u^1, \hdots, u^m).
\end{align}
Without loss of generality, we assume that the goal state is $x^*=\mathbf{0}$.
Finally, we note that through a change of variables, the dynamics of many robotic
systems can be written using polynomials; see, e.g., \cite{posa2015,shen2020}.

\subsection{Control Lyapunov Functions (CLFs)}
A polynomial function $V(x)$ is a \emph{control Lyapunov function} (CLF) 
if it satisfies
the following conditions for some positive integer $\alpha$:
\begin{subequations}
\begin{align}
	&V(x) \ge \epsilon(x^Tx)^\alpha \;\forall x\label{eq:positive_V}\\
	&V(\mathbf{0}) = 0\label{eq:V_at_0}\\
	\begin{split}
	&\text{if } V(x) < \rho \text{ and } x\neq \mathbf{0} \text{ then } \exists u\in\mathcal{U}\\&\qquad\text{ s.t } \underbrace{L_fV+L_gVu}_{\dot{V}(x, u)} < -\kappa_V V \label{eq:Vdot},
\end{split}
\end{align}\label{eq:CLF}
\end{subequations}
where $L_fV=\frac{\partial V}{\partial x}f(x)$  and  $L_gV=\frac{\partial V}{\partial x}g(x)$ 
denote Lie-derivatives.

Under these conditions, the sublevel set $\Omega_\rho=\{x \in \mathbb{R}^{n_x} |V(x)<\rho\}$ 
is an inner approximation of the \emph{stabilizable region}, i.e., the backward reachable
set of the goal state $\mathbf{0}$. This means that for all initial states in $\Omega_\rho$,
there exist control actions that drive the state to $\mathbf{0}$. The
condition~\eqref{eq:positive_V} guarantees that $V(x)$ is positive definite and
radially unbounded, while the condition~\eqref{eq:Vdot} guarantees that $V(x)$
converges to zero exponentially with a rate larger than $\kappa_V > 0$. Our
goal is to find a polynomial CLF $V(x)$ and a scalar $\rho$ that
satisfy~\eqref{eq:CLF} and maximize (in some sense) the size of the $\Omega_\rho$.
  In other words, we seek a CLF that yields a large
\emph{inner} approximation of the stabilizable region. 
We remark that a previous technique for \emph{outer} approximation appears in \cite{majumdar2014convex}.

\subsection{Control Barrier Functions (CBFs)}
Given an unsafe set $\mathcal{X}_{\text{unsafe}}$, a polynomial function $h(x)$
is a \emph{control barrier function} with the safe region $\{x \in \mathbb{R}^{n_x} | h(x)> 0\}$ if it
satisfies \begin{subequations}
\begin{align}
	&h(x) \le 0 \;\forall x\in\mathcal{X}_{\text{unsafe}}\label{eq:cbf_unsafe}\\
	\begin{split}
	&\text{if } h(x)>\beta^-\text{ then } \exists u\in\mathcal{U}\\&\quad\text{ s.t } \underbrace{L_fh + L_ghu}_{\dot{h}(x, u)} > -\kappa_h h\label{eq:cbf_derivative},
\end{split}
\end{align}
\label{eq:CBF}
\end{subequations}
where $\beta^- < 0$ and $\kappa_h>0$ are given constants. We assume that the unsafe region is given as the union of semialgebraic sets, i.e.,
\begin{subequations}
\begin{align}
	\mathcal{X}_{\text{unsafe}}=\mathcal{X}_{\text{unsafe}}^1 \cup\hdots \cup\mathcal{X}_{\text{unsafe}}^{n_{\text{unsafe}}}\\
	\mathcal{X}_{\text{unsafe}}^i=\{ x | p_{i,1}(x)\le 0, \hdots, p_{i, s_i}(x)\le 0\},
\end{align}\label{eq:unsafe_region}
\end{subequations}\noindent where $p_{i, j}(x))$ is a polynomial. Our goal is to find a polynomial CBF $h(x)$ with a large certified safe region $\{x \in \mathbb{R}^{n_x} | h(x)> 0\}$.

Note that the CLF and CBF conditions are related. Specifically, if
$V(x)$ satisfies~\eqref{eq:Vdot} for a given $\kappa$, then
$h(x)=-V(x)$ satisfies~\eqref{eq:cbf_derivative} with the same
$\kappa$ and $\beta^-=-\rho$. Hence any approach for synthesizing CLFs can be used to synthesize
CBFs with slight modification. 

\section{APPROACH}
\label{sec:approach}
In this section, we characterize CLF/CBF functions using SOS conditions,
accounting for input constraints. We then show that these conditions can be verified
by solving an SOS optimization problem. Finally, we present algorithms for automatic
CLF/CBF synthesis that optimize
an inner approximation of the stabilizable or safe region by solving a sequence
of SOS optimization problems.

\subsection{CLF Certification}
\label{subsec:certify_clf}
For brevity, we first focus on CLFs and later show
how these ideas can be modified for CBFs.
To begin, observe that direct use of the CLF condition~\eqref{eq:CLF} is complicated
by equation~\eqref{eq:Vdot}, which, for each state $x\in\Omega_\rho$,
requires existence of an admissible control $u\in\mathcal{U}$ 
satisfying the Lyapunov inequality \eqref{eq:Vdot}.  Current CLF design techniques~\cite{jarvis2003, majumdar2013control}
ensure existence of $u$ by finding an explicit polynomial
control policy satisfying~\eqref{eq:Vdot}. 
This imposes conservatism, since existence of a CLF does not imply existence of a
polynomial controller.  It also adds a semidefinite constraint of order
$\mathcal{O}((\text{degree}(u))^{n_x})$) to the underlying semidefinite
program, limiting scalability. 

We next propose an alternative
strategy based on the contrapositive statement of condition \eqref{eq:Vdot}:
	\begin{equation}
		\begin{split}
		\text{if }\dot{V}(x, u)\ge -\kappa_VV\;\forall u\in\mathcal{U},\\\text{ then } V(x)\ge\rho \text{ or } x=\mathbf{0}.
	\end{split}
	\label{eq:Vdot_contrapositive}
	\end{equation}
Note by taking contrapositive statement, we have replaced the inconvenient
$\exists$ quantifier with the $\forall$ quantifier, which is easier to work
with in an SOS framework. Indeed, as quoted from
\cite[Chapter~9]{tedrakeunderactuated}, \textit{``Our sum-of-squares toolkit is
well-suited for addressing questions with the $\forall$ quantifier over
indeterminates. Working with the $\exists$ quantifier is much more difficult"}. 

\begin{figure}
	\centering
	\includegraphics[width=0.2\textwidth]{\FigureFolder/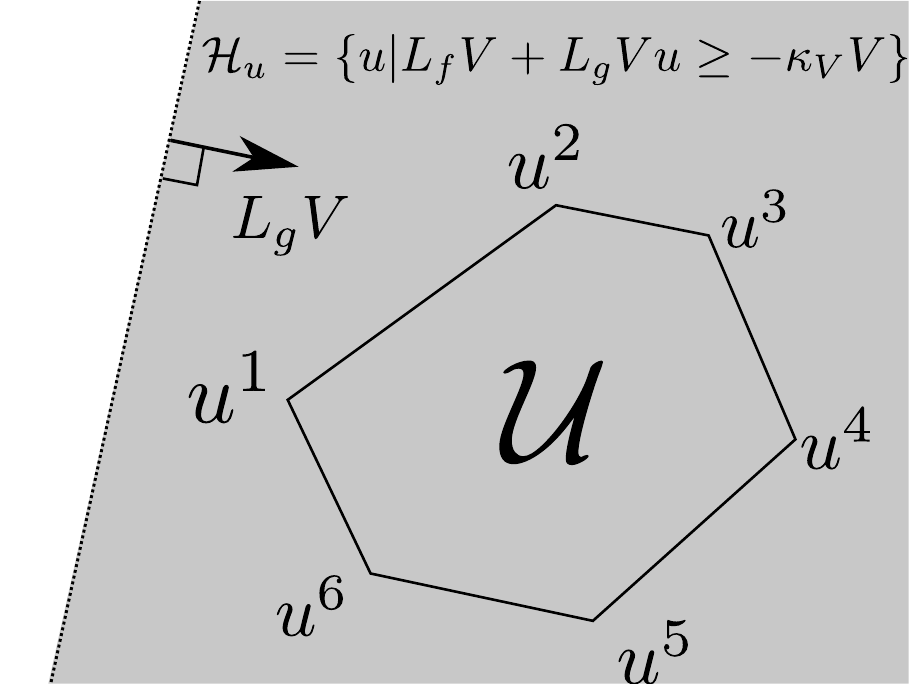}
	\caption{The halfspace $\mathcal{H}_u$ (the shaded region, with the halfspace normal vector as $L_gV$) contains the polytope $\mathcal{U}$ if and only if it contains all vertices of $\mathcal{U}$. This idea is extensively used in robust linear programming \cite{ben2009} and robotics \cite{zeng2021, dai2016}.}
	\label{fig:halfspace_contains_polytope}
\end{figure}

The condition $\dot{V}(x, u)\ge -\kappa_VV\;\forall u\in\mathcal{U}$ 
of the contrapositive \eqref{eq:Vdot_contrapositive} 
denotes infinitely many constraints on $x$ indexed by $u \in \mathcal{U}$.
We next reformulate this using finitely many constraints by exploiting
the polyhedrality of $\mathcal{U}$ and the control-affine dynamics.
To begin, we note that $\dot{V}(x, u) = L_fV + L_gVu$ is a
linear function of $u$ under the control-affine assumption. 
This implies that $\dot{V}(x, u)\ge -\kappa_VV$ is
a linear inequality on $u$ which,  by convexity,
 holds for all $u \in \mathcal{U}$
if and only if it holds at each vertex of $\mathcal{U}$; see Fig. \ref{fig:halfspace_contains_polytope} and also Appendix~\ref{subsec:Vdot_contrapositive_alternate}. 
Hence, $\dot{V}(x, u)\ge -\kappa_VV\;\forall u\in\mathcal{U}$
is equivalent to the finite set of inequalities:
\begin{equation}
\begin{gathered}
	\dot{V}(x, u^i)\ge -\kappa_VV\; \forall i=1,\hdots,m.
\end{gathered}
\label{eq:halfspace_contains_polytope_clf}
\end{equation}
Substituting \eqref{eq:halfspace_contains_polytope_clf} into
\eqref{eq:Vdot_contrapositive}, we arrive at the desired reformulation
of~\eqref{eq:Vdot_contrapositive}:
\begin{equation}
	\begin{split}
	\text{if }\dot{V}(x,u^i)\ge -\kappa_VV\;\forall i=1,\hdots,m,\\
	\text{ then } (V(x)-\rho)x^Tx\ge 0,
\end{split}
\label{eq:Vdot_contrapositive_wo_u}
\end{equation}
where we have also  replaced $V(x)\ge \rho \text{ or } x=\mathbf{0}$ with the
equivalent condition that $(V(x)-\rho)x^Tx\ge 0$.

In summary, we can certify the CLF condition~\eqref{eq:Vdot} by checking if the
polynomial $(V(x)-\rho)x^Tx$ is always non-negative on the semialgebraic set
$\{x | \dot{V}(x, u^i)\ge -\kappa_VV\;i=1,\hdots,m\}$. The
\textit{Positivstellensatz} in turn yields the following necessary and sufficient condition for
non-negativity:
\begin{equation}
	\begin{split}
	\exists q(x), r(x)\in\textbf{preorder}(\dot{V}(x, u^i)+\kappa_VV,i=1,\hdots,m), k\in\mathbb{N}\\ \text{s.t } q(x)(V(x)-\rho)x^Tx = ((V(x)-\rho)x^Tx)^{2k} + r(x).
\end{split}
\label{eq:Vdot_psatz}
\end{equation}
where $\textbf{preorder}(\bullet)$ is the set of polynomials defined in \eqref{eq:preorder}.

Recalling that membership in $\textbf{preorder}(\bullet)$ can be checked with SOS optimization
yields our first key result.
\begin{theorem}
	Given a polynomial $V(x)$ satisfying $V(x)\ge \epsilon (x^Tx)^\alpha$ and $V(\mathbf{0})=0$, $V(x)$ is a valid CLF, with the sublevel set $\Omega_{\rho}=\{x | V(x)<\rho\}$ as an inner approximation of the stabilizable region, if and only if there exists $q(x), r(x), k$ satisfying the convex constraints in \eqref{eq:Vdot_psatz}, certified by the feasibility of the SOS program.
	\label{theorem:clf_psatz}
\end{theorem}

In summary, this theorem shows we can directly certify CLFs using SOS programming.
and, unlike~\cite{jarvis2003, majumdar2013control}, this direct certification
does not require explicit construction of a polynomial stabilizing control law.

\subsection{Synthesizing CLFs}
\label{subsec:search_clf}
Theorem \ref{theorem:clf_psatz} illustrates that, for fixed $V(x)$,
a simple bi-section procedure, involving the sequential solution of
SOS feasibility problems, can maximize $\rho$ and hence the estimate
$\Omega_{\rho}$ of the stabilizable region.  
Searching for $V(x)$, however, is more complicated  since the \textbf{preorder} 
is a non-convex function of $V(x)$. This motivates an alternative sufficient
condition, linear in $V(x)$, that drops product-terms from the \textbf{preorder}. 
\begin{lemma}
	A sufficient condition for \eqref{eq:Vdot_contrapositive_wo_u} is the existence of polynomials $\lambda_i(x),i=0,\hdots,m$ such that
	\begin{subequations}
	\begin{align}
		\begin{split}
			&(1+\lambda_0(x))(V(x)-\rho)x^Tx \\
			&\quad-\sum_{i=1}^m \lambda_i(x)(\dot{V}(x, u^i)+\kappa_VV) \text{ is sos}
		\end{split}\label{eq:clf_vdot_s_procedure1}\\
		&\lambda_i(x)\text{ is sos}. \label{eq:clf_vdot_s_procedure2}
	\end{align}
	\label{eq:clf_vdot_s_procedure}
	\vspace{-10pt}
	\end{subequations}
\end{lemma}

The constraint \eqref{eq:clf_vdot_s_procedure1} is bilinear in $V(x)$ and
$\lambda(x)$. Hence, we can fix one and search for the other
in an alternating fashion using SOS optimization.

To enlarge the sublevel set $\Omega_\rho=\{x | V(x)< \rho\}$, we borrow an
idea from~\cite{tan2004}, which measures the size of the sublevel set
$\Omega_\rho$ with an inner ellipsoid $\mathcal{E}_d=\{x |
(x-x_{\mathcal{E}})^TS_{\mathcal{E}}(x-x_{\mathcal{E}})\le d\} \subset
\text{Cl}(\Omega_\rho)$, where $x_{\mathcal{E}}$ and $S_{\mathcal{E}}$ are
given, and $\text{Cl}(\Omega_\rho)$ is the closure of the set $\Omega_\rho$.
The goal is to expand the ellipsoid $\mathcal{E}_d$ by maximizing $d$.

When $V(x)$ and $\rho$ are fixed, we can find a large inner
ellipsoid $\mathcal{E}_d\subset\text{Cl}(\Omega_{\rho})$ by solving the
following SOS program.
\begin{subequations}
\begin{align}
	\max_{d, s_1(x)}& \; d\\
	\begin{split}
		\text{s.t }& (x-x_{\mathcal{E}})^TS_{\mathcal{E}}(x-x_{\mathcal{E}})-d\\&\qquad\quad - s_1(x)(V(x)-\rho) \text{ is sos}
	\end{split}\label{eq:maximize_inner_ellipsoid_clf1}\\
			&s_1(x) \text{ is sos}\label{eq:maximize_inner_ellipsoid_clf2}.
\end{align}
	\label{eq:maximize_inner_ellipsoid_clf}
\end{subequations}
Constraints~\eqref{eq:maximize_inner_ellipsoid_clf1}-\eqref{eq:maximize_inner_ellipsoid_clf2}
are obtained by applying the S-procedure to the statement
``$x\notin\mathcal{E}_d\text{ if } x\notin\text{Cl}(\Omega_{\rho})$", which is
the contrapositive of $\text{Cl}(\Omega_{\rho})\subset\mathcal{E}_d$.
Keeping $V(x)$ and $\rho$ fixed, we then find the polynomial $\lambda(x)$ in \eqref{eq:clf_vdot_s_procedure} through the following SOS program
\begin{subequations}
\begin{align}
      &\text{find}\;{\lambda(x)}\\
      &\text{subject to constraint } \eqref{eq:clf_vdot_s_procedure}.
\end{align}~\label{eq:search_lagrangian_clf}
\end{subequations}

\begin{figure}
	\begin{subfigure}{0.15\textwidth}
		\includegraphics[width=\textwidth]{\FigureFolder/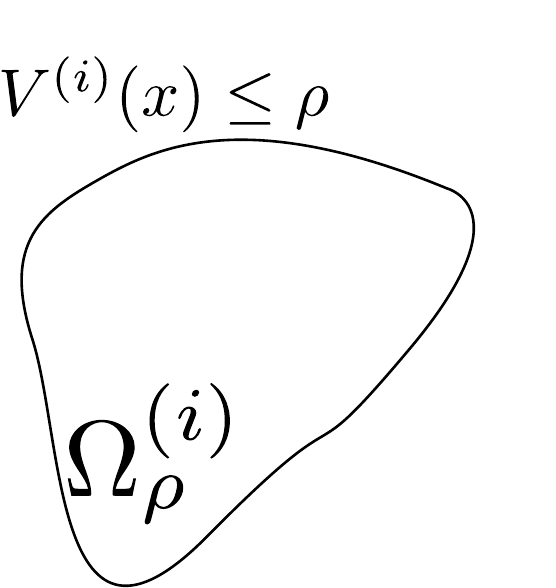}
		\caption{}
	\end{subfigure}
	\begin{subfigure}{0.15\textwidth}
		\includegraphics[width=\textwidth]{\FigureFolder/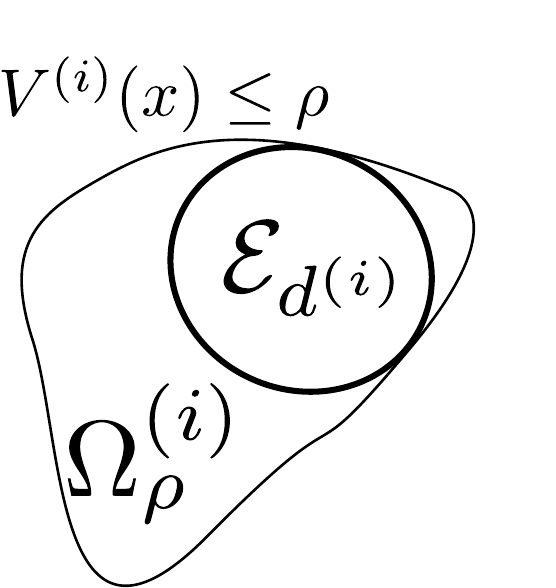}
		\caption{}
	\end{subfigure}
	\begin{subfigure}{0.15\textwidth}
		\includegraphics[width=\textwidth]{\FigureFolder/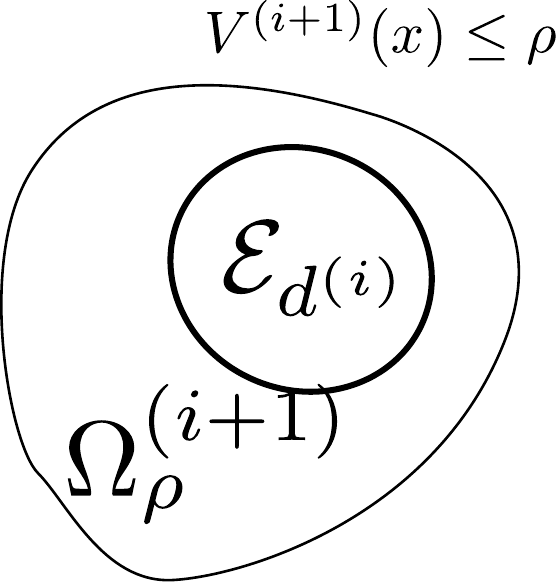}
		\caption{}
	\end{subfigure}
	\caption{(a) At iteration $i$ we have the CLF $V^{(i)}(x)$ with the sublevel set $\Omega^{(i)}_\rho$. (b) We find the largest inner ellipsoid $\mathcal{E}_{d^{(i)}}$ with given ellipsoid center and shape. (3) We update $V^i(x)$ to $V^{(i+1)}(x)$ by increasing the margin between  $\Omega_\rho$ and the ellipsoid $\mathcal{E}_{d^{(i)}}$.}
	\label{fig:maximize_clf_roa_ellipsoid}
\end{figure}

Finally, we update the CLF $V(x)$, aiming to 
 enlarge the sublevel set $\Omega_\rho$ while containing the
ellipsoid $\mathcal{E}_d$. To do this, we increase the margin between the
sublevel set $\Omega_\rho$ and the ellipsoid $\mathcal{E}_d$
by minimizing the maximal value of $V(x)$ on $\mathcal{E}_d$.
Formally, we solve the following optimization
\begin{subequations}
\begin{align}
	\min_{V(x), s_2(x), t}& t \\
	\begin{split}
		\text{s.t }& t - V(x) -s_2(x)(d-(x-x_{\mathcal{E}})^TS_{\mathcal{E}}(x-x_{\mathcal{E}}))\\ &\qquad\text{ is sos}
	\end{split}\label{eq:search_V_given_ellipsoid1}\\
			      &s_2(x)\text{ is sos}\label{eq:search_V_given_ellipsoid2}\\
			      &\text{Constraints } \eqref{eq:positive_V}, \eqref{eq:V_at_0}, \eqref{eq:clf_vdot_s_procedure1},
\end{align}
\label{eq:search_V_given_ellipsoid}
\end{subequations}
where $\lambda(x)$ is fixed to the solution in the program
\eqref{eq:search_lagrangian_clf}. Constraints
\eqref{eq:search_V_given_ellipsoid1}-\eqref{eq:search_V_given_ellipsoid2}
guarantee that $t\ge \max_{x\in\mathcal{E}_d}V(x)$.
We present our algorithm in Algorithm \ref{algorithm:clf_ellipsoid}, and visualize it pictorially in Fig. \ref{fig:maximize_clf_roa_ellipsoid}.

\begin{algorithm}
	\caption{Finding CLF with inscribed ellipsoids through bilinear alternation}
	\label{algorithm:clf_ellipsoid}
	\begin{algorithmic}
		\State Start with $V^{(0)}(x), i=0, \text{converged=False}$
		\While{not converged}
		\State Solve SOS program \eqref{eq:maximize_inner_ellipsoid_clf} to find the ellipsoid $\mathcal{E}_{d^{(i)}}$.
		\If{$d^{(i)} \le d^{(i-1)}$}
		\State converged=True
		\Else
		\State Solve SOS program \eqref{eq:search_lagrangian_clf} to find the polynomials $\lambda^{(i)}(x)$.
		\State Solve SOS program \eqref{eq:search_V_given_ellipsoid} to find the CLF $V^{(i+1)}$, $i=i+1$.
		\EndIf
		\EndWhile
	\end{algorithmic}
\end{algorithm}

The goal of Algorithm~\ref{algorithm:clf_ellipsoid} is to maximize 
the size of $\Omega_\rho$. An alternative goal is to stabilize
a \emph{specified set} of initial conditions.  To this end, we ensure
that $\Omega_\rho$ contains a set of specified states $x^{(j)}, j=1,\hdots,
n_{\text{sample}}$ by minimizing the maximal of $V(x^{(j)})$ 
on this set:
\begin{subequations}
\begin{align}
	&\min_{V(x), \lambda(x)} \max_{j=1,\hdots,n_{\text{sample}}}V(x^{(j)})\label{eq:search_V_minimize_sample1}\\
	&\text{subject to constraint } \eqref{eq:clf_vdot_s_procedure}
\end{align}
\label{eq:search_V_minimize_sample}
\end{subequations}
This optimization program~\eqref{eq:search_V_minimize_sample} has the bilinear
product between $\lambda(x)$ and $V(x)$. We solve it with bilinear
alternations using Algorithm \ref{algorithm:clf_sample}.
\begin{algorithm}
	\caption{Find CLF by minimizing sample values with bilinear alternation}
	\label{algorithm:clf_sample}
	\begin{algorithmic}
		\State Start with $V(x^{(0)}), i=0$, converged=False
		\While{not converged}
		\State Fix $V^{(i)}(x)$, find $\lambda^{(i)}(x)$ through solving the SOS program \eqref{eq:search_V_minimize_sample}.
		\State Fix $\lambda^{(i)}(x)$, find $V^{(i+1)}(x)$ through solving the SOS program \eqref{eq:search_V_minimize_sample}. Denote the objective value as $o^{(i+1)}$
		\If{$o^{(i)}-o^{(i+1)} < tol$}
		\State converged=True
		\EndIf
		\EndWhile
	\end{algorithmic}
\end{algorithm}

After we find $V(x)$ through either Algorithm \ref{algorithm:clf_ellipsoid} or
\ref{algorithm:clf_sample}, we can then further enlarge the sublevel set
$\Omega_\rho$ using bisection on $\rho$ to satisfy the necessary and sufficient
condition \eqref{eq:Vdot_psatz}.

\subsection{Extension to CBFs}~\label{subsec:cbf}
Similar to CLFs, we can synthesize CBFs by solving SOS programs
using arguments identical to those from \ref{subsec:certify_clf}. 
We sketch the details here and defer details
to the appendix. To begin, we note that the  contrapositive
statement of \eqref{eq:cbf_derivative} is
	\begin{equation}
		\text{if } \dot{h}(x, u) \le -\kappa_hh\;\forall u\in\mathcal{U},\text{ then } h(x) \le \beta^-, \label{eq:cbf_derivative_contrapositive1}
      \end{equation}
which, same as for CLFs, can be rewritten using finitely many inequalities.
Specifically, it is equivalent to
	\begin{equation}
		\text{if } \dot{h}(x, u^i) \le -\kappa_hh\;\forall i=1,\hdots,m, \text{ then } h(x)\le \beta^-
	\label{eq:cbf_derivative_contrapositive},
      \end{equation}
a consequence of  the control-affine dynamics and polyhedrality of $\mathcal{U}$
as described in~\ref{subsec:certify_clf}.

The CBF condition requires non-negativity of $\beta^- - h(x)$ 
on the semialgebraic set $\{x | \dot{h}(x,
u^i)\le-\kappa_hh,\;i=1,\hdots,m\}$, which, by the \textit{Positivstellensatz}
can be reformulated as a  SOS program,  as done for CLFs in \ref{subsec:certify_clf}. To
search for a CBF while expanding the certified safe region, we can formulate
the sufficient condition as an SOS program and solve a sequence of SOS programs,
similar to the algorithms for CLFs in \ref{subsec:search_clf}. We
present the detailed mathematical formulation and algorithms for certifying and
searching CBFs in Appendix \ref{subsec:certify_cbf} and
\ref{subsec:search_cbf}\footnote{Contemporarily, \cite{zhao2022} derived
a similar SOS-based formulation for verifying another safety certificate called \textit{safety index}, and solved the SOS program through nonlinear optimization instead of convex optimization.}.

\section{Results}
We show our results on a 2D toy system, an inverted pendulum and a 3D
quadrotor. We use Mosek \cite{andersen2000mosek} to solve the SOS optimization
problems arising in our synthesis procedures.
\subsection{2D toy system}
\begin{figure}
	\centering
	\includegraphics[width=0.4\textwidth]{\FigureFolder/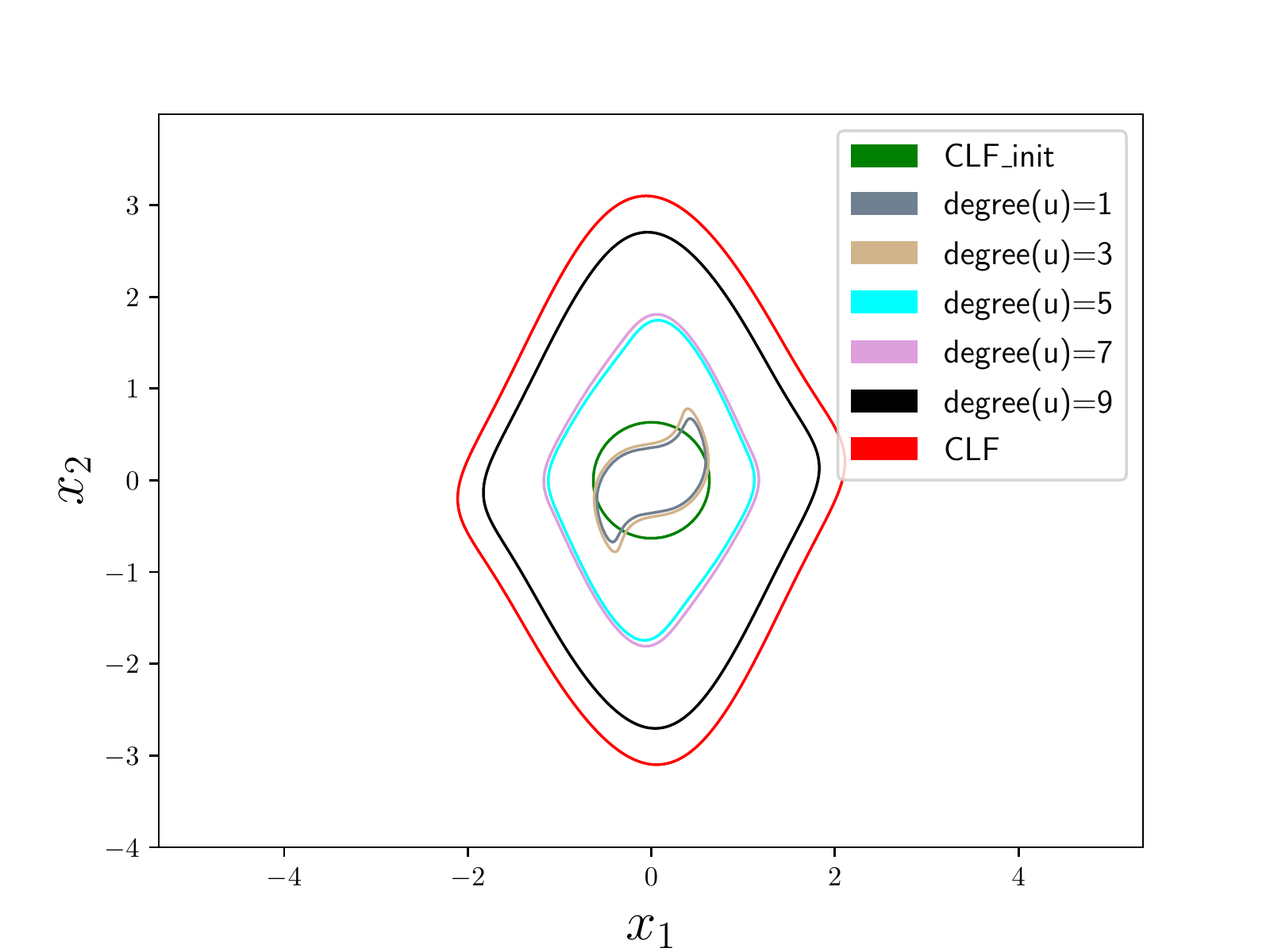}
	\caption{We draw the certified inner approximation of the stabilizable region for the 2D dynamical system (Eq.\eqref{eq:tan_dynamics}). Starting from the certified region with the innermost circle, our algorithm expands the sublevel set $\Omega_\rho$ to the outermost curve. We also compare with the certified inner approximation of the stabilizable region by searching over a polynomial controller, and draw the region for each controller degree.}
	\label{fig:packard_V_contours}
\end{figure}

Our first example is a 2D toy system from \cite{tan2004},
given by
\begin{subequations}
\begin{align}
	\dot{x}_1 =& u\\
	\dot{x}_2 =& -x_1+\frac{1}{6}x_1^3-u,
\end{align}
\label{eq:tan_dynamics}
\end{subequations}
with input constraint $-0.4\le u\le 0.4$. 
Using Algorithm~\ref{algorithm:clf_ellipsoid}, we certify a stabilizable region
$\Omega_{\rho} = \{x | V(x)\le\rho\}$ using a CLF $V(x)$ of degree 8,
initializing with $V(x) = x_1^2 + x_2^2$ and $\rho = 0.3$.  We maximize the
final $\rho$ using  bisection and Theorem~\ref{theorem:clf_psatz}.  The initial
and final sublevel sets are plotted as the inner-most green curve and
outer-most red curve in Fig. \ref{fig:packard_V_contours}.  As illustrated, our
algorithm greatly increases the size of the certified region. The computation
time for each SOS program is less than 0.01s.

An alternative approach to certify  $\Omega_{\rho}$
proceeds by explicitly searching for a polynomial controller $u(x)$
satisfying $\dot V(x, u(x)) < 0$ and the input limits.
Using the formulation proposed in \cite{jarvis2003} (and also explained in the
Appendix \ref{subsec:search_V_and_u}), we plot the certified
regions obtained for controllers of increasing degree in Fig. \ref{fig:packard_V_contours}.
As shown, for the same $V(x)$, our CLF approach certifies a larger $\rho$, as it does not restrict
to polynomial control laws. Rather, it allows any stabilizing control action within the input limits.
This demonstrates the advantage of our approach versus jointly
searching for a polynomial Lyapunov function and a polynomial controller.

\subsection{Inverted pendulum}
\begin{figure}
	\centering
	\includegraphics[width=0.12\textwidth]{\FigureFolder/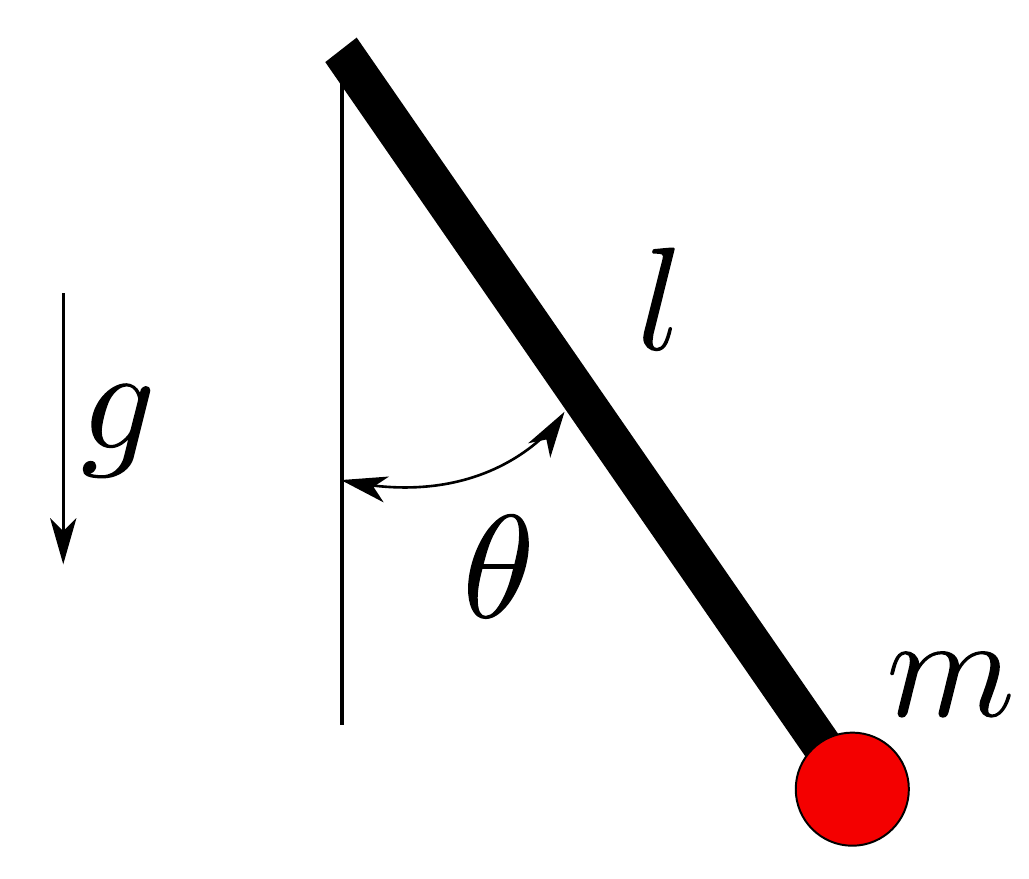}
	\caption{Pendulum}
	\label{fig:pendulum}
\end{figure}

For the inverted pendulum (Fig. \ref{fig:pendulum}) described in
\cite{tedrakeunderactuated} (with mass $m=1\text{kg}$, length $l=0.5\text{m}$
and damping $b=0.1\text{N}/\text{m}$, we aim to stabilize the upright
equilibrium $\theta=\pi, \dot{\theta}=0$ using input limits $-4.6\text{N}\cdot
\text{m} \le u \le 4.6 \text{N}\cdot \text{m}$).
Following~\cite{wampler2011, posa2015, shen2020}, we formulate the dynamics
as polynomials of the state-vector $x = [s, c+1,
\dot{\theta}]$ where $s=\sin\theta, c=\cos\theta$. Specifically, we take 
\begin{align}
	\dot{x} = \begin{bmatrix} (x_2-1)x_3\\-x_1x_3\\-\frac{mglx_1+bx_3}{ml^2}\end{bmatrix} + \begin{bmatrix}0 \\ 0 \\ \frac{1}{ml^2}\end{bmatrix}u,
\end{align}
and impose the additional algebraic constraint that $x_1^2 + (x_2-1)^2=1$
(since $\sin^2\theta+\cos^2\theta=1$). We incorporate this constraint into the SOS
programs using the S-procedure~\cite{shen2020, posa2015}.

\begin{figure}
	\begin{subfigure}{0.28\textwidth}
	\includegraphics[width=0.98\textwidth]{\FigureFolder/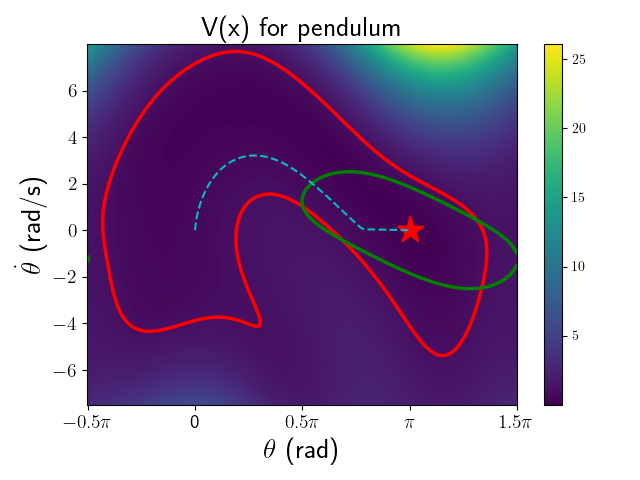}
	\caption{}
	\label{fig:pendulum_V}
\end{subfigure}
\begin{subfigure}{0.2\textwidth}
	\includegraphics[width=0.98\textwidth]{\FigureFolder/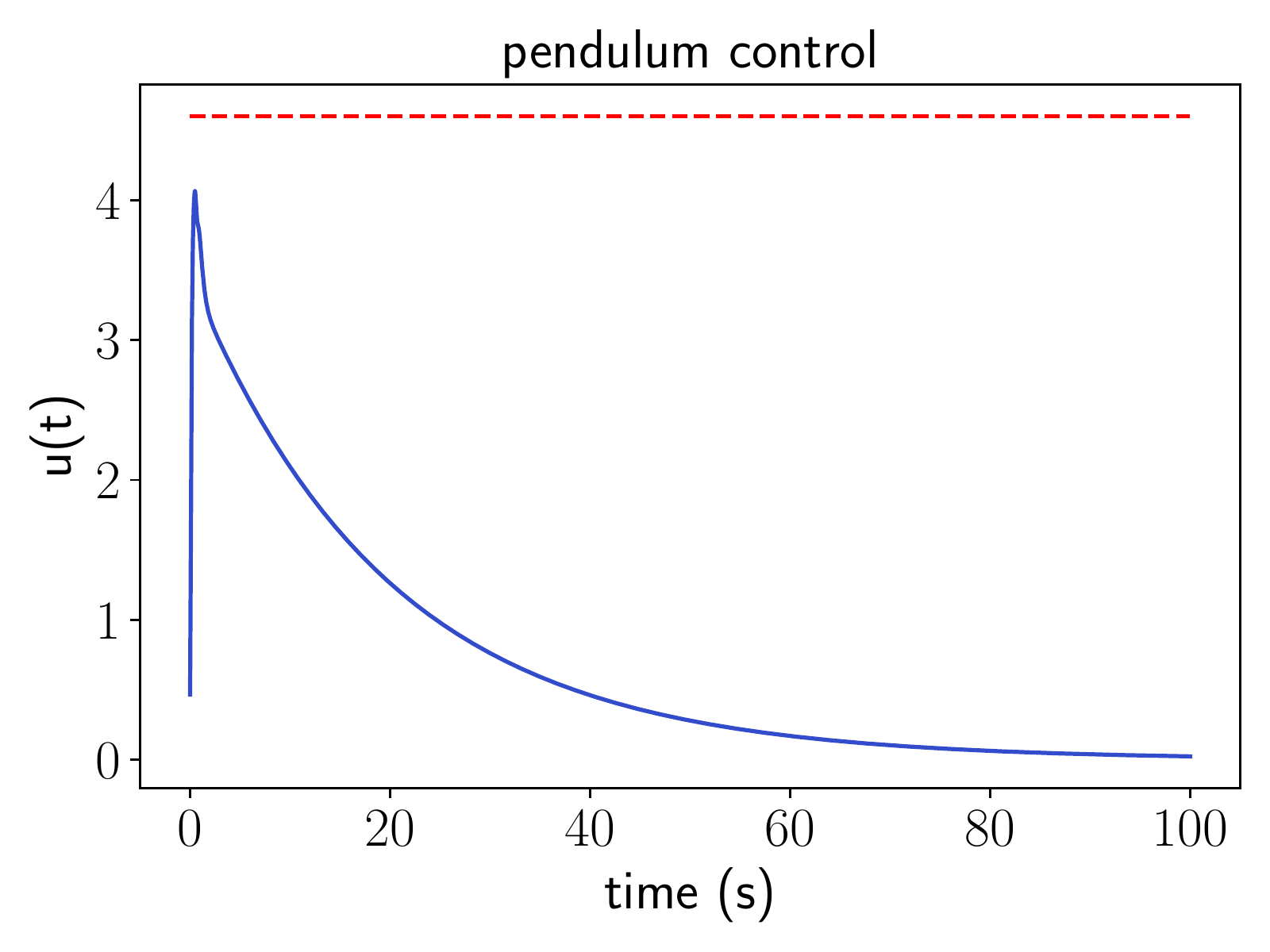}
	\caption{}
	\label{fig:pendulum_control_time}
\end{subfigure}
\caption{(Left) The heatmap of the CLF found by Algorithm \ref{algorithm:clf_sample}. The red contour is the boundary of the sublevel set $\{x | V(x)\le\rho\}$. The green countour is the boundary of initial certified stabilizable region. We draw the simulated trajectory from $\theta=0, \dot{\theta}=0$ as the cyan dashed line. (Right) The control action along the simulated trajectory. The red dashed line is the torque limit.}
\vspace{-12pt}
\end{figure}

Initializing the CLF to the LQR quadratic cost-to-go function (green contour in Fig.
\ref{fig:pendulum_V}), we apply Algorithm \ref{algorithm:clf_sample} 
to cover the downright equilibrium state $(\theta, \dot{\theta}) = (0, 0)$ with 
the certified sublevel set $\Omega_\rho$. For a 4-th degree $V(x)$,
we  plot the final $\Omega_\rho$ as the red contour in Fig. \ref{fig:pendulum_V}.
We simulate the pendulum using a QP controller derived from the CLF.
Specifically, we minimize $u^2$ subject to the constraint $\dot{V}(x, u)\le -\kappa_VV,
-4.6\le u\le 4.6$. In Fig. \ref{fig:pendulum_control_time} we plot the torque
values from this QP-based controller on this simulated trajectory. The
computation time for each SOS program in Algorithm
\ref{algorithm:clf_sample} is about 0.15s.

\subsection{Quadrotor}
We apply our approach to a 3D quadrotor to demonstrate its scalability. The system has 13 states (with the orientation represented by a unit quaternion $z$), whose dynamics can be written as polynomial functions of states \cite{fresk2013full} with the additional algebraic constraint $z^Tz=1$ on the unit quaternion. We impose the input limit constraint as $0\le u \le 0.75mg$ for each rotor thrust.

To find a CLF, we initialize Algorithm \ref{algorithm:clf_ellipsoid} with the 
quadratic LQR cost-to-go function. For a quadrotor with body length$=0.15\text{m}$, our certified
stabilizable region covers distant initial states including $p_{xyz}=(10, 0,
0), \text{RollPitchYaw}=(100^\circ, 0, 0), \text{vel}=0$, and the state
$p_{xyz}=(5, 5, 0), \text{RolPitchYaw}=(140^\circ, 0, 0), \text{vel}=0$ (shown
in Fig.  \ref{fig:quadrotor_clf_traj}). Each SOS program in Algorithm
\ref{algorithm:clf_ellipsoid} takes about 615 seconds. We compare with the
computation time of jointly searching for a Lyapunov function and a polynomial
controller (as explained in Appendix \ref{subsec:search_V_and_u}). 
Finding a linear controller takes 11 seconds, and a cubic controller 1540 seconds.
 Synthesis of a 5-th degree controller fails due to insufficient
memory on our 128 GB machine. Computation time grows rapidly with the controller
degree given that the SOS program has an underlying semidefinite constraint of order
$\mathcal{O}((\text{degree}(u))^{13})$ for a 13-state robot. In constrast,
our approach avoids this scalability issue, since it does not require
explicit construction of a polynomial control law.

\begin{figure}
	\begin{subfigure}{0.25\textwidth}
	\includegraphics[width=0.98\textwidth]{\FigureFolder/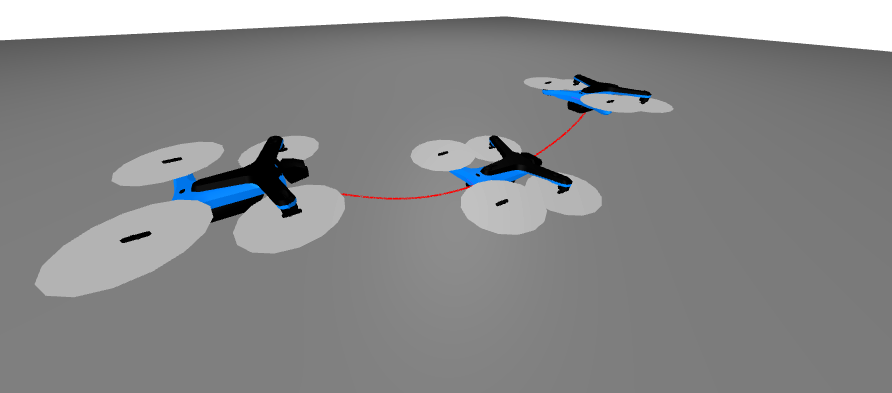}
	\caption{}
	\label{fig:quadrotor_traj1}
\end{subfigure}
\begin{subfigure}{0.23\textwidth}
	\includegraphics[width=0.98\textwidth]{\FigureFolder/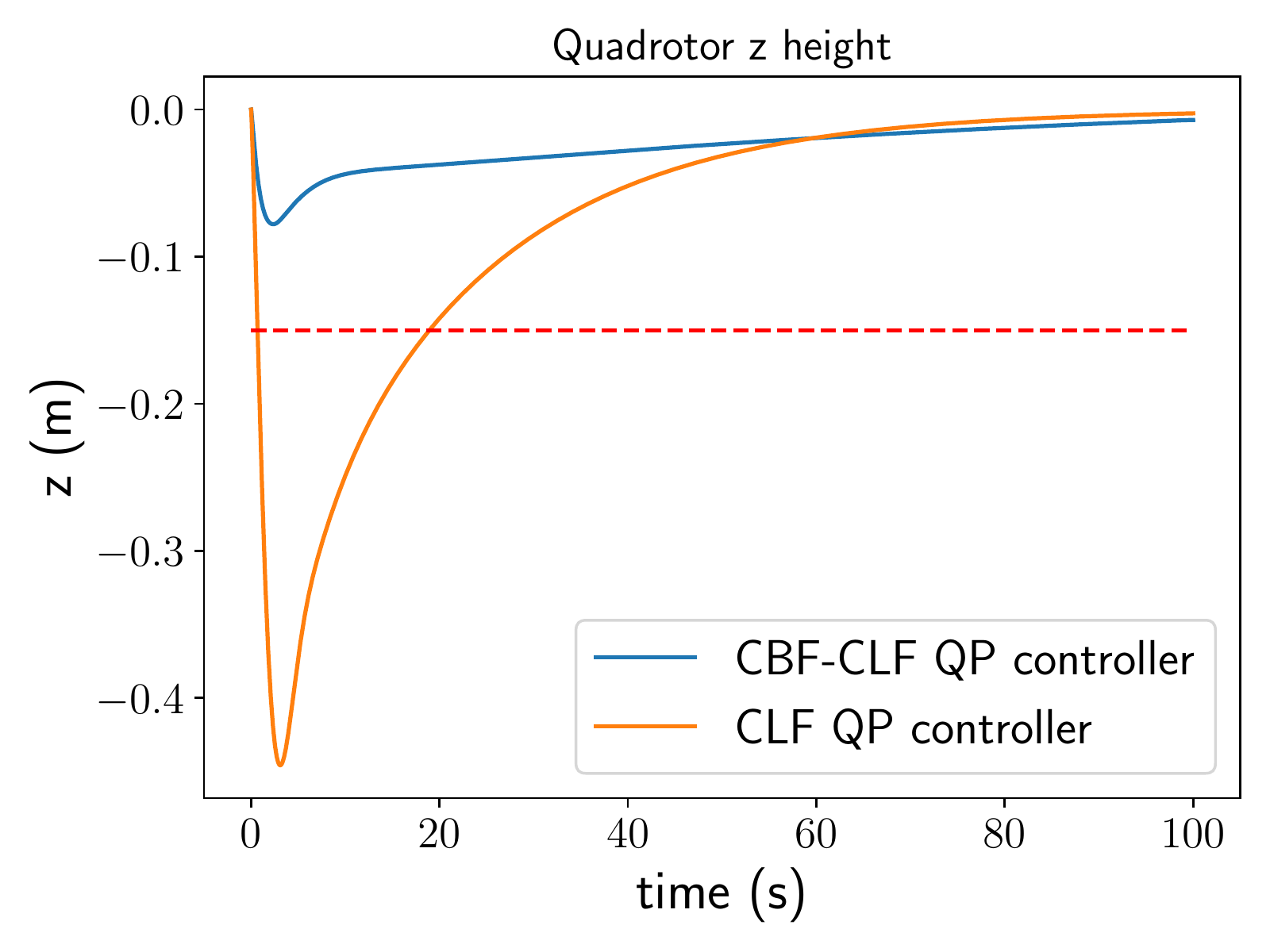}
	\caption{}
	\label{fig:quadrotor_clf_cbf_z_1}
\end{subfigure}
\caption{(a) Simulated quadrotor above grey ground with CBF-CLF-QP controller, flying from right to the left. (b) quadrotor z height along the simulated trajectories. The red dashed line is the boundary of the unsafe set $\{x\in\mathbb{R}^{13} | p_z \le -0.15m\}$.}
\vspace{-12pt}
\end{figure}

For this example, we also synthesize a controller that respects a
minimum height constraint. Specifically, we synthesis a CBF 
for the unsafe set $\mathcal{X}_{\text{unsafe}}=\{x | p_z \le -0.15\}$. 
We start with the initial CBF as $0.0001 - V(x)$ and expand the certified safe region using
Algorithms \ref{algorithm:cbf1} and \ref{algorithm:cbf2}. We simulate the
quadrotor with the CBF-CLF-QP controller \cite{ames2019control} which considers
both CBF and CLF constraints, and compare with the CLF-QP controller that does
not take the unsafe region into consideration. We visualize the simulated
trajectory (starting from hovering at $p_x=1m, p_y=p_z=0m$) in Fig.
\ref{fig:quadrotor_traj1}. We compare the simulation result with the CBF-CLF-QP
controller versus with the CLF-QP controller in Fig.
\ref{fig:quadrotor_clf_cbf_z_1}. Without the CBF constraint the quadrotor drops
below $p_z=-0.15m$ into the unsafe set. With the CBF constraint it always stays
within the safe region. 

%\begin{figure}
%	\begin{subfigure}{0.24\textwidth}
%		\centering
%		\includegraphics[width=0.98\textwidth]{\FigureFolder/quadrotor_clf_cbf_V_1.pdf}
%	\end{subfigure}
%	\begin{subfigure}{0.24\textwidth}
%		\centering
%		\includegraphics[width=0.98\textwidth]{\FigureFolder/quadrotor_clf_cbf_h_1.pdf}
%	\end{subfigure}
%	\caption{Quadrotor CLF and CBF value along the simulated trajectory}
%	\label{fig:quadrotor_clf_cbf_1}
%\end{figure}

\section{Conclusion and discussion}
\label{sec:conclusion}
In this paper we characterized polynomial CLF/CBF functions for
input-constrained, polynomial dynamical-systems using SOS conditions. We then
posed CLF/CBF synthesis as an SOS program whose solution provides an inner
approximation of the stabilizable/safe region. We presented iterative
algorithms to expand the inner approximation of the stabilizable/safe
region using sequential SOS optimization.  Finally, we showcased our approach
on different dynamical systems and compared with explicitly searching for
polynomial controllers using Lyapunov functions.

Currently,  our approach applies to systems with continuous dynamics. Future
work could extend it to  hybrid systems with rigid contacts by
leveraging ideas from~\cite{posa2015}.

\bibliography{acc}
\bibliographystyle{plain}

\section{APPENDIX}
\subsection{Alternative derivation for Eq.\eqref{eq:Vdot_contrapositive_wo_u}}
\label{subsec:Vdot_contrapositive_alternate}
We present an alternative approach to derive \eqref{eq:Vdot_contrapositive_wo_u} from \eqref{eq:Vdot}. Note that \eqref{eq:Vdot} can be written as
\begin{align}
	\text{if } V(x)<\rho\text{ and } x\neq\mathbf{0}\text{ then } \min_{u\in\mathcal{U}}\dot{V}(x, u)< -\kappa_VV\label{eq:Vdot_min}
\end{align}
Since $\dot{V}(x, u)$ is linear in $u$, the minimization of this linear function over the polytope $\mathcal{U}$ has to occur at one of the polytope vertices. Hence condition \eqref{eq:Vdot_min} is equivalent to
\begin{align}
	\text{if } V(x)<\rho\text{ and } x\neq\mathbf{0}\text{ then } \min_{i=1,\hdots,m}\dot{V}(x, u^i)< -\kappa_VV\label{eq:Vdot_min_vertices}
\end{align}
The contrapositive statement of \eqref{eq:Vdot_min_vertices} is 
\begin{align}
	\text{if }\min_{i=1,\hdots,m}\dot{V}(x, u^i)\ge-\kappa_VV \text{ then } V(x)\ge \rho\text{ or } x=\mathbf{0},
\end{align}
which is equivalent to \eqref{eq:Vdot_contrapositive_wo_u}.

\subsection{CBF certification}
\label{subsec:certify_cbf}
Similar to CLFs in section \ref{subsec:certify_clf}, we apply \textit{Positivstellensatz} on equation \eqref{eq:cbf_derivative_contrapositive} to certify the CBF condition. Given a polynomial $h(x)$ satisfying $h(x)\le 0 \;\forall x\in\mathcal{X}_{\text{unsafe}}$, the necessary and sufficient condition for $h(x)$ being a CBF is
\begin{equation}
	\begin{split}
	\exists q_h(x), r_h(x)\in\textbf{preorder}(-\dot{h}(x, u^i)-\kappa_hh,i=1,\hdots,m)\\ k_h\in\mathbb{N}, \text{ s.t } q_h(x)(\beta^--h) = (\beta^--h)^{2k_h}+r_h(x).
\end{split}
	\label{eq:cbf_psatz}
\end{equation}
$h(x)$ is a valid CBF if the SOS program \eqref{eq:cbf_psatz} is feasible.

\subsection{Search certifiable CBFs}
\label{subsec:search_cbf}
Similar to the CLF case, this necessary and sufficient condition is applicable when we certify whether a given $h(x)$ is a valid CBF, but not suitable to search for $h(x)$ due to the nonlinear product of $h(x)$ in the \textbf{preorder}. Instead we present the following sufficient condition of \eqref{eq:cbf_derivative_contrapositive} which is linear in $h(x)$.
\begin{lemma}
	A sufficient condition for the CBF condition \eqref{eq:cbf_derivative_contrapositive} (and \eqref{eq:cbf_derivative}) is the existence of polynomials $\mu_i(x), i=0,\hdots,m$, such that
	\begin{subequations}
	\begin{align}
		\begin{split}
		&(1+\mu_0(x))(\beta^--h(x))\\&\qquad+\sum_{i=1}^m\mu_i(x)(\dot{h}(x, u^i)+\kappa_hh) \text{ is sos}
		\end{split}\label{eq:cbf_derivative_sos1}\\
		&\mu_i(x) \text{ is sos}, i=0,\hdots,m\label{eq:cbf_derivative_sos2}.
	\end{align}
	\label{eq:cbf_derivative_sos}
\end{subequations}
\end{lemma}

A sufficient condition for $h(x)\le 0\;\forall x\in\mathcal{X}_{\text{unsafe}}$ (Equation \eqref{eq:cbf_unsafe}) is the existence of polynomials $\phi(x)$ such that
\begin{subequations}
\begin{align}
	-(1+\phi_{i,0}(x))h(x) + \sum_{j=1}^{s_i}\phi_{i, j}p_{i, j}(x)\text{ is sos}\label{eq:cbf_unsafe_sos1}\\
	\phi_{i, j} \text{ is sos} \;i=1,\hdots,n_{\text{unsafe}}, j=0,\hdots,s_i\label{eq:cbf_unsafe_sos2},
\end{align}
\label{eq:cbf_unsafe_sos}
\end{subequations}
where the polynomials $p_{i, j}(x)$ define the unsafe region $\mathcal{X}_{\text{unsafe}}$ as unions of semialgebraic sets (Equation \eqref{eq:unsafe_region}).

In order to find a CBF $h(x)$ that certifies a large safe region $\Phi=\{x | h(x)>0\}$, we again measure the size of the superlevel set $\Phi$ with the inscribed ellipsoid $\mathcal{E}_d = \{x | (x-x_{\mathcal{E}})^TS_{\mathcal{E}}(x-x_{\mathcal{E}})\le d\}$ given $x_{\mathcal{E}}$ and $S_{\mathcal{E}}$. We find a large $d$ such that $\mathcal{E}_d\subset\text{Cl}(\Phi)$ through the following SOS program 
\begin{subequations}
\begin{align}
	\max_{d, \psi(x)} d\\
	\text{s.t }(x-x_{\mathcal{E}})^TS_{\mathcal{E}}(x-x_{\mathcal{E}}))-d - \psi(x)h(x)\text{ is sos}\\
	\psi(x)\text{ is sos}.
\end{align}
\label{eq:maximize_inner_ellipsoid_cbf}
\end{subequations}

With a given $h(x)$, we can certify if it is a valid CBF by searching for $\mu(x), \phi(x)$ through the following SOS program
\begin{subequations}
\begin{align}
	\text{find } \mu(x),\phi(x)\\
	\text{subject to constraint } \eqref{eq:cbf_derivative_sos} \text{ and }\eqref{eq:cbf_unsafe_sos}.
\end{align}
\label{eq:cbf_search_lagrangian}
\end{subequations}

With $\mu(x),\phi(x)$ obtained through program \eqref{eq:cbf_search_lagrangian}, and the ellipsoid $\mathcal{E}_d$ from program \eqref{eq:maximize_inner_ellipsoid_cbf}, we can search for $h(x)$ with the aim of enlarging the margin between the certified safe region $\Phi=\{x|h(x)>0\}$ and the ellipsoid $\mathcal{E}_d$, we  maximize the minimal value of $h(x)$ over the ellipsoid $\mathcal{E}_d$
\begin{subequations}
\begin{align}
	&\max_{t, h(x), \nu(x)} t\\
	\begin{split}
		\text{s.t }& h(x)-t - \\&\qquad\nu(x)(d-(x-x_{\mathcal{E}})^TS_{\mathcal{E}}(x-x_{\mathcal{E}})\text{ is sos}
\end{split}\\
	&\nu(x) \text{ is sos}\\
	&\text{constraint } \eqref{eq:cbf_derivative_sos1} \text{ and } \eqref{eq:cbf_unsafe_sos1}\\
	& h(x_{\text{anchor}}) \le 1\label{eq:h_at_anchor},
\end{align}
\label{eq:cbf_search_h}
\end{subequations}
where $x_{\text{anchor}}$ is a given safe state. We impose constraint \eqref{eq:h_at_anchor} to prevent scaling $h(x)$ with a infinitely large factor.

We present our algorithm to iterative search for CBF $h(x)$ while enlarging the safe region in Algorithm \ref{algorithm:cbf1}.
\begin{algorithm}
	\caption{Search CBFs through bilinear alternation with inner ellipsoid}
	\label{algorithm:cbf1}
	\begin{algorithmic}
		\State Start with $h^{(0)}(x), i=0$, converged=False
		\While{not converged}
		\State Fix $h^{(i)}$, solve SOS program \eqref{eq:maximize_inner_ellipsoid_cbf} to find $d^{(i)}$.
		\If{$d^{(i)} - d^{(i-1)}<tol$}
		\State converged = True
		\Else
		\State Fix $h^{(i)}(x)$, solve SOS program \eqref{eq:cbf_search_lagrangian} to find $\mu^{(i)}(x),\psi^{(i)}(x)$.
		\State Fix $\mu^{(i)}(x), \psi^{(i)}(x)$, solve SOS program \eqref{eq:cbf_search_h} to find $h^{(i+1)}$, $i=i+1$.
		\EndIf
		\EndWhile
	\end{algorithmic}
\end{algorithm}

Similar to the CLF case, we can also search for $h(x)$ with the aim of certifying some given states $x^{(k)}, k=1,\hdots,n_{\text{sample}}$. The certified safe region $\Phi=\{x | h(x)>0\}$ covers the sampled state if $h(x^{(k)})>0$. So our goal is to maximize the minimal value of $h(x^{(k)}), k=1,\hdots,n_{\text{sample}}$ so as to bring them all positive through the following program
\begin{subequations}
\begin{align}
	&\max_{h(x), \mu(x), \phi(x)} \min_{k=1,\hdots,n_{\text{sample}}}h(x^{(k)})\label{eq:cbf_sample1}\\
	&\text{subject to constraint } \eqref{eq:cbf_derivative_sos} \eqref{eq:cbf_unsafe_sos}\eqref{eq:h_at_anchor} \label{eq:cbf_sample2}.
\end{align}
\label{eq:cbf_sample}
\end{subequations}
Problem \eqref{eq:cbf_sample} has the bilinear product of $h(x)$ and $\mu(x), \phi(x)$, we present our bilinear alternation algorithm in Algorithm \ref{algorithm:cbf2}.
\begin{algorithm}
	\caption{Search CBFs through bilinear alternation with sampled states}
	\label{algorithm:cbf2}
	\begin{algorithmic}
		\State Given $h^{(0)}(x), i=0$ and converged=False
		\While{not converged}
		\State Fix $h^{(i)}(x)$, search $\mu^{(i)}(x),\phi^{(i)}(x)$ satisfying constraints \eqref{eq:cbf_sample2}.
		\State Fix $\mu^{(i)}(x), \phi^{(i)}(x)$, search $h^{(i+1)}(x)$ through the SOS program \eqref{eq:cbf_sample}. Denote the objective value as $o^{(i+1)}$.
		\If{$o^{(i+1)}-o^{(i)}< tol$}
		\State converged=True
		\EndIf
		\EndWhile
	\end{algorithmic}
\end{algorithm}

\subsection{Formulation for searching a polynomial controller and $\rho$}
\label{subsec:search_V_and_u}
Given a polynomial function $V(x)$ that satisfies $V(x)\ge \epsilon(x^Tx)^{\alpha}, V(\mathbf{0})=0$, to verify that this $V(x)$ is a valid Lyapunov function with polynomial control policy $u(x)$ and inner approximation of region-of-attraction $\{x | V(x)\le\rho\}$, we impose the following constraint
\begin{subequations}
\begin{align}
	L_fV + L_gVu(x)\le-\kappa_VV\text{ if } V(x)\le\rho\\
	-u_{\text{lo}} \le u(x)\le u_{\text{up}} \text{ if } V(x)\le\rho.
\end{align}
\label{eq:search_V_and_u}
\end{subequations}
Using S-procedure, a sufficient condition for \eqref{eq:search_V_and_u} is that the following SOS program is feasible.
\begin{subequations}
\begin{align}
	\text{Find} u(x), \gamma(x), \eta_{\text{lo}}(x), \eta_{\text{up}}(x)\\
	-\kappa_VV-L_fV-l_gVu(x) - \gamma(x)(\rho-V(x))\text{ is sos}\\
	u_{\text{up}} - u - \eta_{\text{up}}(x)(\rho - V(x))\text{ is sos}\\
	u - u_{\text{lo}} - \eta_{\text{lo}}(x)(\rho - V(x))\text{ is sos}\\
	\gamma(x), \eta_{\text{up}}(x), \eta_{\text{lo}}(x) \text{ are sos}.
\end{align}
\label{eq:search_V_and_u_sos}
\end{subequations}
We can then find the maximal $\rho$ through bisection with SOS program \eqref{eq:search_V_and_u_sos}.

\end{document}